\documentclass[11pt,a4paper]{article}
\usepackage[hyperref]{acl2019}
\usepackage{times}
\usepackage{latexsym}
\usepackage{comment}
\usepackage{amsmath}
\usepackage{makecell}
\usepackage{amssymb}
\usepackage{times}  
\usepackage{xcolor}
\usepackage{helvet}  
\usepackage{courier}  
\usepackage{url}  
\usepackage{graphicx}  
\usepackage[autostyle]{csquotes}
\usepackage{url}

\newcommand{\x}[0]{\mathbf{x}}
\newcommand{\h}[0]{\mathbf{h}}
\newcommand{\vv}[0]{\mathbf{v}}
\newcommand{\bb}[0]{\mathbf{b}}
\newcommand{\s}[0]{\mathbf{s}}
\newcommand{\p}[0]{\mathbf{p}}
\newcommand{\cc}[0]{\mathbf{c}}
\newcommand{\softmax}[0]{\text{softmax}}
\newcommand{\I}[0]{\mathcal{I}}

\aclfinalcopy 

\title{Simple, Fast, Accurate Intent Classification and Slot Labeling for Goal-Oriented Dialogue Systems}

\author{Arshit Gupta\thanks{\ \ Equal Contribution} \\
  Amazon AI \\
  Seattle \\
  {\tt arshig@amazon.com} \\\And
  John Hewitt\footnotemark[1] \hspace{0.005em} \thanks{\ \ Work performed while at Amazon AI} \\
  Stanford University  \\
  Palo Alto\\
  {\tt johnhew@stanford.edu} \\\And
  Katrin Kirchhoff \\
  Amazon AI  \\
  Seattle \\
  {\tt katrinki@amazon.com} \\}

\date{}

\begin{document}
\maketitle
\begin{abstract}
  
  With the advent of conversational assistants, like Amazon Alexa, Google Now, etc., dialogue systems are gaining a lot of traction, especially in industrial setting. 
  These systems typically consist of Spoken Language understanding component which, in turn, consists of two tasks - Intent Classification (IC) and Slot Labeling (SL). Generally, these two tasks are modeled together jointly to achieve best performance. However, this joint modeling adds to model obfuscation. 
  In this work, we first design framework for a modularization of joint IC-SL task to enhance architecture transparency. 
 Then, we explore a number of self-attention, convolutional, and recurrent models, contributing a large-scale analysis of modeling paradigms for IC+SL across two datasets.  Finally, using this framework, we propose a class of ‘label-recurrent’ models that otherwise non-recurrent, with a 10-dimensional representation of the label history, and show that our proposed systems are easy to interpret, highly accurate (achieving over 30\% error reduction in SL over the state-of-the-art on the Snips dataset), as well as fast, at 2x the inference and 2/3 to 1/2 the training time of comparable recurrent models, thus giving an edge in critical real-world systems.
\end{abstract}

\section{Introduction}

\begin{figure}
  \centering
  \includegraphics[width=.75\linewidth]{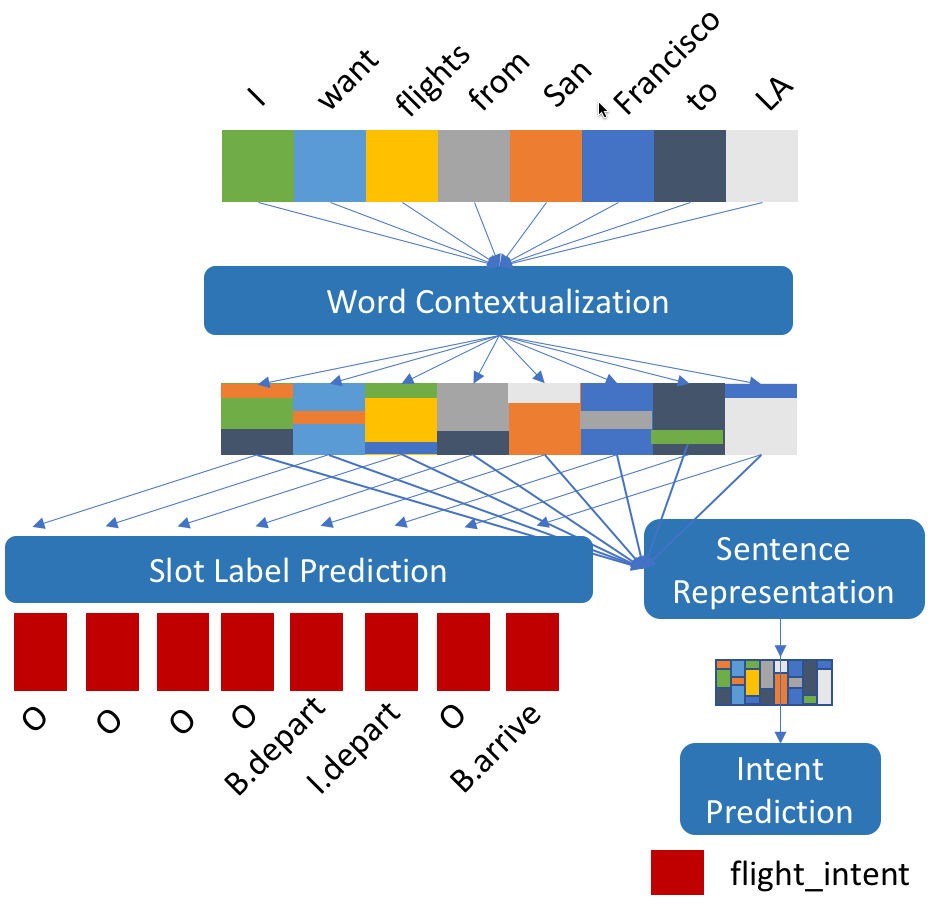}
  \caption{\small A general framework of joint IC+SL, decoupling modeling tasks to permit the analysis of each component independently.}
  \label{joint_framework_diagram}
\end{figure}

At the core of task-oriented dialogue systems are spoken language understanding (SLU) models, tasked with determining the intent of users’ utterances and labeling semantically relevant words at each turn of the conversation. Performance on these tasks, known as intent classification (IC) and slot labeling (SL), upper-bounds the utility of such dialogue systems. A large body of recent research has improved these models through the use of recurrent neural networks, encoder-decoder architectures, and attention mechanisms. However, for production dialogue systems in particular, system speed is at a premium, both during training and in real-time inference.

In this work, we propose fully non-recurrent and label-recurrent model paradigms including self-attention and convolution for comparison to state-of-the-art recurrent models in terms of accuracy and speed.
To achieve this, we design a framework for joint IC-SL models that is modularized into different components and makes the task agnostic to type of neural network used. 
This, in turn, makes the model architecture simpler, easy to understand and renders the task network agnostic, allowing for easier plug and play using existing components, such as pre-trained contextual word embeddings \cite{devlin2018bert}, etc. This is \emph{essential} for easier model debugging and quicker experimentation, especially in industrial setting.


Using this framework, we identify three distinct model families of interest: fully recurrent, label-recurrent, and non-recurrent.
Recent state-of-the-art models fall into the first category, as encoder-decoder architectures have recurrent encoders to perform word context encoding, and predict slot label sequences using recurrent decoders that use both word and label information as they decode \cite{hakkani2016multi,liu16b,li2018self}. In second category, we have `non-recurrent' networks: fully feed-forward, attention-based, or convolutional models, for example. Lastly, we have a class of label-recurrent models, inspired by structured sequential models like conditional random fields on top of non-recurrent word contextualization components. In this class of models, slot label decoding proceeds such that label sequences are encoded by a recurrent component, but word sequences are not.

Our contributions are:
\begin{itemize}
\item A class of label-recurrent convolutional models that achieve state-of-the-art performance on Snips and competitive performance on ATIS while maintaining faster training and inference speeds than fully-recurrent models
\item A new modular framework for joint IC-SL models that permits the analysis of individual modeling components that decomposes these joint models into separate components for \textit{word context encoding}, \textit{summarization of the sentence} into a single vector for intent classification, and \textit{modeling of dependencies in the output space} of slot label sequences. 
\item In-depth analysis of different word contextualizations for Spoken Language Understanding task (for instance, providing evidence for the intuition that explicitly focusing on local context is a useful architectural inductive prior for slot labeling) 
\end{itemize} 





\section{Prior Work}
There is a large body of research in applying recurrent modeling advances to intent classification and slot labeling (frequently called spoken language understanding). Traditionally, for intent classification, word n-grams were used with SVM classifier \cite{haffner2003optimizing} and Adaboost \cite{schapire2000boostexter}. For the SL task, CRFs \cite{gorin1997may} have been used in the past.

Recently, a larger focus has been on joint modeling of IC and SL tasks. Long short-term memory recurrent neural networks \cite{hochreiter1997long} and Gated Recurrent Unit models \cite{cho2014learning} were proposed for slot labeling by \citet{yao2014spoken} and \citet{zhang2016joint} respectively, while \citet{guo14} used recursive neural networks.
Subsequent improvements to recurrent neural modeling techniques, like bidirectionality and attention \cite{Bahdanau2014NeuralMT} were incorporated into IC+SL in recent years as well \cite{hakkani2016multi,liu16b}. \citet{li2018self} introduced a self-attention based joint model where they used self-attention and LSTM layers along with the gating mechanism for this task. 

Non-recurrent modeling for language has been re-visited recently, even as recurrent techniques continue to be dominant.
Dilated CNNs \cite{yu2015multi} with CRF label modeling were applied to named entity recognition by \citet{strubell17}, and earlier were applied to SL by \citet{xu13}.
Convolutional and attention-based sentence encoders have been applied in complex tasks, including machine translation, natural language inference, and parsing. \cite{gehring2017convolutional,vaswani2017attention,shen17,kitaev18constituency}
We draw from both of these bodies of work to propose a simple yet highly effective family of IC+SL models.

\section{A general framework of joint IC+SL}
Intent classification and slot labeling take as input an utterance $\x_{1:T} = \{\x_1, \x_2, ... \x_T\}$, composed of words $\x_i$ and of length $T$.
Models construct a distribution over intents and slot label sequences given the utterance.
One intent is assigned per utterance and one slot label is assigned per word:
\begin{align}
  P(l_{1:T},c \ | \x_{1:T})
\end{align}
where $c \in \I$, a fixed set of intents, and $l_i \in \mathcal{L}$, a fixed set of slot labels.
Models are trained to minimize the cross-entropy loss between the assigned distribution and the training data.
To the end of constructing this distribution, our framework explicitly separates the following components, which are explicitly or implicitly present in all joint IC+SL systems (Figure \ref{joint_framework_diagram}):

\subsection{Word contextualization}
We first assume words are encoded through an embedding layer, providing context-independent word vectors.
Overloading notation, we denote the embedded sequence $\x_{1:T}$, with $\x_i \in \mathbb{R}^{d_x}$. 

    In this component, word representations are enriched with sentential context. Each word $\x_{i}$ is assigned a contextualized representation $\h_{i}$.
    To ease layering these components, we keep the dimensionality the same as the word embeddings; $\h_{i} \in \mathbb{R}^{d_x}$.
Our study consists mainly of varying this component across models which are described in detail in Section \ref{word_context}.
In all models, we assume independence of intent classification and slot labeling given the learned representations:
\begin{align}
  P(l_{1:T},c|\h_{1:T}) = P(l_{1:T}|\h_{1:T})P(c|\h_{1:T})
\end{align}

  \subsection{Sentence representation} In this component, the output of the word contextualization component is summarized in a single vector,
    \begin{align}
      \s = \text{SentenceRepr}(\h_{1:T})
    \end{align}
    where $\s \in \mathbb{R}^{d_x}$. 
    For all our experiments, we keep this component constant, using a simple attention-like pooling which is the weighted sum of word contextualization for each position in the sentence. These weights are computed using softmax over these individual word contextualizations
    
    While simple, this model permits word contextualization components freedom in how they encode sentential information; for example, self-attention models may spread full-sentence information across all words, whereas 1-directional LSTMs may focus full-sentence information in the last word's vector.

\subsection{Intent prediction}

   In this component, the sentence representation is used as features to predict the intent of the utterance.
   For all experiments, we keep this component fixed as well, using a simple  two-layer feed-forward block on top of $\mathbf{s}$.

\subsection{Slot label prediction}
   In this component, the output of the word contextualization component is used to construct a distribution over slot label sequences for the utterance.
    We decompose the joint probability of the label sequence given the contextualized word representations into a left-to-right labeling:
    \begin{align}
      P(l_{1:T}|\h_{1:T}) = \prod_{i=1}^{T}P(l_i|\h_{1:T},l_{1:i-1})
    \end{align}
    In our experiments, we explore two models for slot prediction, one fully-parallelizable because of strong independence assumptions, the other permitting a constrained dependence between labeling decisions that we call `label-recurrent'.

\paragraph{Independent slot prediction}
    The first is a non-recurrent model, which assumes indepdencence
    between all labeling decisions once given $\h_{1:T}$, as well as independence from all word representations except that of the word being labeled:
    \begin{align}
      P(l_i|\h_{1:T},l_{1:i-1}) = P(l_i|\h_{i})
    \end{align}
    This model is fully parallelizable on GPU architectures, and the probability of each labeling decision is modeled according to
    \begin{align}
      P(l_i|\h_{1:T}) =&\; \softmax(W^{(3)}\p_{i,\text{SL}} + \bb^{(3)})\\
      \p_{i,\text{SL}} =&\; \text{tanh}(W^{(4)}\h_i + \bb^{(4)}) \label{eqn_p_sl}
    \end{align}
    hence, SL prediction features are learned using each contextualized word independently.
    
       \paragraph{Label-recurrent slot prediction}
    The second class of slot prediction models we consider lead to our classification, `label-recurrent.'\footnote{We use this term for clarity in language, not to claim that no such models have been explored in the past.}
    These models permit dependence of labeling decisions on the sequence of decisions made so far, but keep the independence assumption on the word representations:
    \begin{align}
      P(l_i|\h_{1:T},l_{1:i-1}) = P(l_i|l_{1:i-1},\h_{i})
    \end{align}
    Notably, this family of models excludes traditional encoder-decoder models, since the decoder component uses labeling decisions $l_{1:i-1}$ and earlier word representations $\h_{1:i-1}$ to influence the predictor features $\p_{i,SL}$.
    However, it includes models such as CNN-CRF.
    
       The space of label sequences in slot labeling is much smaller than the space of word sequences. This adds minimal computational burden and permits the model to benefit from GPU parallelism during $\h_{1:T}$ computation.
    
    For our experiments, we propose a single label-recurrent model, which encodes labeling histories $l_{1:-i}$ using only a 10-dimensional LSTM.
    First, slot labels are embedded, such that for each $l \in \mathcal{L}$, we have $\mathbf{l} \in \mathbb{R}^{d_l}$.
    An initial tag history state, $h_{0}^{\text{tag}}$, is randomly initialized.
    Each tag decision is fed along with the previous tag history state to the LSTM, which returns the next tag history state:
    \begin{align}
      \h_{i}^{\text{tag}} = \text{LSTM}(\mathbf{l}_{i},\h^{\text{tag}}_{i-1}).
    \end{align}
    We omit a precise description of the LSTM model here for space, referring the reader to \cite{hochreiter1997long}.

    The tag history is used at each prediction step as additional inputs to construct the predictor features $\p_{i,\text{SL}}$, replacing Eqn.~\ref{eqn_p_sl} with:
  \begin{align}
    \p_{i,\text{SL}} =&\; \text{tanh}(W^{(5)}[\h_i;\h_i^{\text{tag}}] + \bb^{(5)}) 
  \end{align}
  where $[a;b]$ denotes concatenation.
  This model and other label-recurrent models are not only parallelizable more than fully-recurrent models, but also provide an architectural inductive bias, separating modeling of tag sequences from modeling of word sequences.
	In our experiments, we perform greedy decoding to maintain a high decoding speed.

\section{Word contextualization models} \label{word_context}
In this section, we describe word contextualization models with the goal of identifying non-recurrent architectures that achieve high accuracy and faster speed than recurrent models.

\subsection{Feed-forward model}
In this model, we set $\h_{1:T} = \x_{1:T} + \mathbf{a}_{1:T}$, where $\mathbf{a}_{1:T}$ is a learned absolute position representation, with one vector learned per absolute position, as used in \cite{gehring2017convolutional}.
While extremely simple, this model provides a useful baseline as a totally context-free model.
It also permits us to analyze the contribution of a label-recurrent component in such a context-deprived scenario.

\subsection{Self-attention models}
Recent work in non-recurrent modeling has surfaced a number of variants of attention-based word context modeling.

The simplest constructs each $\h_i$ by incorporating a weighted average of the rest of the sequence, $\x_{1:T}\backslash \x_i$.
We use a general bilinear attention mechanism with a residual connection while masking out the identity in the attention weights.
\begin{align}
  \h_i =&\; \text{relu}(\sqrt{.5}(\cc_i + \x_i)) \\ 
  \cc_i =&\; \sum_{j=1, j\not=i}^{T}\alpha_{j}\x_{j}\\
  \alpha_{j} =&\; \frac{\exp(\x_i^TW^{(5)}\x_{j})}{\sum_{j'=1}^{T}\exp({\x_i^TW^{(5)}\x_{j'})}}
\end{align}
In this and all subsequent models, we optionally stack multiple layers, feeding the word representations from each layer into the next; in this case we denote the models \textsc{attn-1l}, \textsc{attn-2l}, etc.

We also analyze multi-head attention models, drawing from \cite{vaswani2017attention}. For a model with $k$ heads, we construct one matrix of the form $A\in\mathbb{R}^{d_x/k}$ for each head, and transform each $\x_i$, $\x_{i}^{k'} = A^{k'}\x_{i}$ for $k' \in \{1,...,k\}$.
These are passed into the attention equations above, generating context vectors ${\cc_i^{1},...,\cc_{i}^{k}} \in \mathbb{R}^{d_x/k}$, which are then concatenated to form a vector in $\mathbb{R}^{d_x}$.
These context layers are usually sent through a linear transformation to combine features between the heads, the output of which is $\cc_i$, but we found that omitting this combination transformation leads to significantly improved results, so we do so in all experiments.
We denote these models \textsc{k-head attn}.

\subsubsection{Relative position representations}
We found in early experiments that the absolute position embeddings in self-attention models are insufficient for representing order. Hence, in all attention models except when explicitly noted, we use relative position representations as follows.
We follow \citet{shaw2018self}, who improved the absolute position representations of the Transformer model \cite{vaswani2017attention} by learning vector representations of relative positions and incorporating them into the self-attention mechanism as follows:
\begin{align}
  \cc_i =&\; \sum_{i'=1, j\not=i}^{T}\alpha_{j}(\x_{j} + \vv_{f(i,j)})\\
  \alpha_{j} =&\; \frac{\exp(\x_i^TW^{(5)}\x_{j} + b_{f(i,j)})}{\sum_{j'=1}^{T}\exp(\x_i^TW^{(5)}\x_{j'} + b_{f(i,j)})}
\end{align}
where $\vv_{f(i,j)}$ is a learned vector representing how the relative positions $i$ and $j$ should be incorporated, and $b_{f(i,j)}$ is a learned bias that determines how the relative position should affect the weight given to position $j$ when contextualizing position $i$.
The function $f$ determines which relative positions to group together with a single relative position vector.
Given the generally small datasets in IC+SL, we use the following relative position function, which buckets relative positions together in exponentially larger groups as distance increases, following the results of \citet{khandelwal2018lm}, that LSTMs represent position fuzzily at long relative distances.
\begin{align}
  \small
  f(i,j) = \begin{cases}
              \pm 1\, , |j-i| = 1\\
              \pm 2\, , |j-i| \in \{2,3\}\\
              \pm 3\, , |j-i| \in \{4..7\}\\
              ...
           \end{cases}
\end{align}
This is similar to the very recent preprint work of \citet{bilan2018position}, who use linearly increasing bucket sizes; we found exponentially increasing sizes to work well compared to the constant bucket sizes of \citet{shaw2018self}.

\subsection{Convolutional models}
Convolution incorporates local word context into word representations, where kernel width parameter specifies the total size (in words) of local context considered.
Each convolutional layer produces a vector representation of each word,
\begin{align}
  \h_{1:T} = \text{relu}(\sqrt{.5}*[\text{CNN}(\x_{1:T}) + \x_{1:T}])
\end{align}
and includes a residual connection, and variance normalization, following \cite{gehring2017convolutional}.
To maintain the dimensionality of $\h_i$ as $\mathbb{R}^{d_x}$, we use a filter count of $d_x$.
We vary the number of CNN layers as well as the kernel width, and for all models use a variant known as dilated CNNs.
These CNNs incorporate distant context into word representations by skipping an increasing number of nearby words in each subsequent convolutional pass.
We use an exponentially increasing dilation size; in the first layer, words of distance 1 are incorporated; at layer two, words of distance 2, then 4, etc.
This permits large contexts to be incorporated into word representations while keeping kernel sizes and the number of layers low.

\subsection{Recurrent models}

We also construct a recurrent word contextualization model, more or less identical to encoders of recent state-of-the-art models.
We use a bidirectional LSTM to encode word contexts, $\h_{1:T} = \text{BiLSTM}(\x_{1:T})$.
As with all other models, we report the performance of this model with feed-forward slot label prediction as well as with label-recurrent slot label prediction.
Though similar to earlier work, both models are new; though the latter is recurrent both in word contextualization and slot label prediction, it is distinct from past models in that the two recurrent components are completely decoupled until the prediction step.

\section{Datasets}
We evaluate our framework and models on the ATIS data set \cite{hemphill90} of
spoken airline reservation requests and the Snips NLU Benchmark
set \cite{coucke2018snips}.
The ATIS training set contains 4978 utterances from the ATIS-2 and ATIS-3
corpora; the test set consists of 893 utterances from the ATIS-3
NOV93 and DEC94 data sets. The number of slot labels is 127, and the
number of intent classes is 18. Only the words themselves are used as
input; no additional tags are used.

\begin{table*}
\centering
  \small

  \begin{tabular}{|l|r|r|r|r|r|r|r|r|r|}
    \hline
    Model & \thead{label\\recurrent} & \multicolumn{2}{c|}{IC acc} & \multicolumn{2}{c|}{SL F1} & \thead{Inference\\ms/utterance} & \thead{Epochs to\\converge} & s/epoch & \#\\
     &  & Snips & ATIS & Snips & ATIS &  &  &  & \\

\hline
\textsc{feed-forward} & No & \bf98.56 & 97.14 & 53.59 & 69.68 & 0.61 & 48 & 1.82 & 17k\\
\hline
\textsc{feed-forward} & Yes & 98.54 & \bf97.46 & \bf75.35 & \bf88.72 & 1.82 & 83 & 2.52 & 19k\\
\hline
\hline
\textsc{cnn, 5kernel, 1l} & No & 98.56 & 98.40 & 85.88 & 94.11 & 0.82 & 23 & 1.90 & 42k\\
\textsc{cnn, 5kernel, 3l} & No & 99.04 & \bf98.42 & 92.21 & 96.68 & 1.37 & 55 & 2.16 & 91k\\
\textsc{cnn, 3kernel, 4l} & No & 98.81 & 98.32 & 91.65 & 96.75 & 1.28 & 57 & 2.29 & 76k\\
\hline
\textsc{cnn, 5kernel, 1l} & Yes & 98.85 & 98.36 & 93.12 & 96.39 & 2.13 & 51 & 2.77 & 43k\\
\textsc{cnn, 5kernel, 3l} & Yes & \bf99.10 & 98.36 & \bf94.22 & \bf96.95 & 2.68 & 59 & 3.34 & 93k\\
\textsc{cnn, 3kernel, 4l} & Yes & 98.96 & 98.32 & 93.71 & \bf96.95 & 2.60 & 53 & 3.43 & 78k\\
\hline
\hline
\textsc{attn, 1head, 1l, no-pos} & No & 98.50 & 97.51 & 53.61 & 69.31 & 1.95 & 25 & 1.94 & 22k\\
\textsc{attn, 1head, 1l} & No & 98.53 & 97.74 & 75.55 & 93.22 & 4.75 & 117 & 4.34 & 23k\\
\textsc{attn, 1head, 3l} & No & \bf98.74 & 98.10 & 81.51 & 94.07 & 7.68 & 160 & 4.32 & 33k\\
\textsc{attn, 2head, 3l} & No & 98.31 & 98.10 & 83.02 & 94.61 & 7.86 & 79 & 4.87 & 47k\\
\hline
\textsc{attn, 1head, 1l, no pos} & Yes & 98.63 & 97.68 & 74.94 & 88.60 & 3.24 & 60 & 2.66 & 24k\\
\textsc{attn, 1head, 1l} & Yes & 98.61 & 98.00 & 86.72 & 94.53 & 6.12 & 89 & 5.53 & 24k\\
\textsc{attn, 1head, 3l} & Yes & 98.51 & \bf98.26 & 88.04 & 94.99 & 9.03 & 109 & 6.06 & 34k\\
\textsc{attn, 2head, 3l} & Yes & 98.48 & \bf98.26 & \bf89.31 & \bf95.86 & 9.17 & 93 & 6.54 & 49k\\
\hline
\hline
\textsc{lstm, 1l} & No & \bf98.82 & 98.34 & 91.83 & 97.28 & 2.65 & 45 & 2.91 & 47k\\
\textsc{lstm, 2l} & No & 98.77 & 98.20 & 93.10 & 97.36 & 4.72 & 58 & 5.09 & 77k\\
\hline
\textsc{lstm, 1l} & Yes & 98.68 & \bf98.36 & 93.83 & \bf97.37 & 3.98 & 54 & 4.62 & 49k\\
\textsc{lstm, 2l} & Yes & 98.71 & 98.30 &\bf93.88 & 97.28 & 6.03 & 69 & 6.82 & 79k\\
\hline

  \end{tabular}
    \caption{\label{table_dev}\small Development results on the Snips 2017 and ATIS datasets, comparing models from feed-forward, convolutional, self-attention, and recurrent paradigms, as well as comparing non-recurrent, label-recurrent, and fully recurrent architectures, on IC, SL, inference speed, and training time.
  Inference speed, convergence time, and parameter count are drawn from Snips experiments, but the trends hold on ATIS.
      The best IC and SL for each dataset is bolded within each model paradigm to help compare between paradigms.}
\end{table*}

The Snips 2017 dataset is a collection of 16K crowdsourced queries, with 
about 2400 utterances per each of 7 intents. These intents range from
`Play Music' to `Get Weather'. Training data contains 13784 utterances and the
 test data consists of 700 utterances. The utterance tokens are mixed case unlike the ATIS
  dataset, where all the tokens are lowercased. Total number of slot labels are 72.
  We use IOB tagging, and split 10\% of the train set off to form a development set.
Utterances in Snips are, on average, short, with 9.15 words per utterance compared to ATIS' 11.2.
However, slot label sequences themselves are longer in Snips, averaging 1.8 tokens per span to ATIS' 1.2, making span-level slot labeling more difficult.
For our development experiments, we use the casing and tokenization provided by Snips. Co, but to compare to prior work, in one test experiment we use the lowercased, tokenized version of \cite{goo2018slot}\footnote{https://github.com/MiuLab/SlotGated-SLU}.


\section{Experiments}
We evaluate multiple models from each of our model paradigms to help determine what modeling structures are necessary for SLU, and where the best accuracy-speed tradeoffs are.
First, we report extensive evaluation across the Snips and ATIS development sets, tracking inference speed and time to convergence along with the usual IC accuracy and SL F1.
Second, we pick a small number of our best-performing models to evaluate on ATIS and Snips test sets, to compare against prior work.

For each experiment below, we train until convergence, where convergence is defined by an early stopping criterion with a patience of 30 epochs and an average of development set IC accuracy and token-level SL F1 used as the performance metric.

\subsection{Modeling study experiments}
In our first category of experiments, we evaluate variants of each word contextualization paradigm introduced.


We evaluate one feed-forward word contextualization module (labeled as \textsc{feed-forward}) to provide a baseline performance.
As with all subsequent models, we evaluate this word contextualization module with and without our proposed label-recurrent decoder.
This baseline should help us determine the extent to which each dataset requires the modeling of context.

We evaluate 3 convolutional word contextualization modules.
The first has 1 layer with a kernel size of 5, and is intended to provide intuition as to whether a relatively large local context can sufficiently model SL behavior.
We label this model \textsc{cnn, 5kernel, 1l}, and name all other CNN models similarly.
The next model has 3 layers with kernel size 5, and is dilated.
This model incorporates long-distance context hierarchically, and is shorter and wider-per-layer than the otherwise-similar 3rd CNN model, with 4 layers and kernel size 3.

We evaluate 4 attention-based word contextualization modules.
The first is simple, with 1 attention head and 1 layer. Unlike all others we analyze, it does not use relative position embeddings.
Thus, this model is word order-invariant except for a simple absolute position embedding.
If it improves over \textsc{feedforward}, then, it provides strong evidence that semantic information from the context words, irrespective of order, is useful in making tagging decisions.
We label this model with the flag \textsc{no-pos}.
To evaluate the utility of relative position embeddings, we also compare a model with 1 head and one layer, labeled \textsc{attn, 1head, 1l}. 
We then test two increasingly complex models, first with 3 layers and 1 head, the second with 3 layers and 2 heads per layer.

We evaluate 2 LSTM-based word contextualization modules; one uses a single LSTM layer, whereas the other stacks a second on top of the first.
As with all other models, we test these two models both with independent slot prediction and label-recurrent slot prediction.


\subsection{Comparison to prior work}
For our second category of experiments, we take a few high-performing models from our analysis and evaluate them on the Snips and ATIS test sets for comparison to prior work.
For these models, we report not only the average IC accuracy and SL F1 across random initializations, but also the standard deviation and best model, as most work has not reported average values.
We keep all hyperparameters fixed across all experiments, potentially hindering performance but providing a stronger analysis of robustness.

\textbf{Note on pre-trained contextual word embedding}: Although our framework easy integration of contextual pre-trained embeddings, like BERT \cite{devlin2018bert} and EMLo \cite{Gardner2017AllenNLP} by replacing the word contextualization component, however, in order to reduce model obfuscation and to have fair comparison against baselines, we exclude them in our experimentation.

\section{Results and discussion}
In this section, we draw from results reported in Table~\ref{table_dev}, on the development sets of Snips and ATIS.
It is easy to see that very little in the way of modeling is necessary for IC task, so we focus our analysis on SL task.
We emphasize that ATIS has shorter spans than Snips, averaging 1.2 and 1.8 tokens, respectively, leading to differing modeling requirements.

\subsection{Minimal modeling for SLU}
By analyzing three simple models - \textsc{feed-forward}, \textsc{attn-1head-1l-no-pos}, and \textsc{cnn-5kernel-1l} - we conclude that explicitly incorporating local features is a useful inductive bias for high SL accuracy.
The purely feed-forward model achieves 53.59 SL F1 on Snips, whereas one layer of convolution improves that number to 85.88.
The story is similar for ATIS SL.
However, a single layer of attention without position information fails to improve over the feed-forward model whatsoever which we believe is due to the order-invariant nature of self-attention.
This also emphasizes the fact that focusing on local context is useful inductive prior for SL task.

For each of these simple models, switching from independent slot label prediction to label-recurrent prediction provides large gains on both datasets. We find an approximate 1.3ms/utterance slowdown from using label recurrence across all models.
Thus, in terms of accuracy-for-speed, very simple models can achieve much of the results of more expensive models as long as they are label-recurrent and incorporate local context.
\begin{table*}
\centering
\small
\begin{tabular}{|l|c|c|c|c|c|}
  \hline
  & & \multicolumn{4}{c|}{Snips}\\
  \hline
  &  &\multicolumn{2}{c|}{IC Acc} & \multicolumn{2}{c|}{SLR F1}\\
  Model & Recurrence & Mean & Max & Mean & Max  \\
  \hline
  16 LSTM* \cite{hakkani2016multi}& full & 96.9 & - & 87.3 & - \\
  '16 seq2seq+attn* \cite{liu16b}& full & 96.7 & - & 87.8 & - \\
  LSTM+attn+gates \cite{goo2018slot}& full & 97.0 & - & 88.8 & - \\
  \hline
	\textsc{Our CNN, 5Kernel, 3L} & none & \bf97.65$\pm$0.28 & 97.57 & 89.57$\pm$0.54 & 90.66\\
	\textsc{Our CNN, 5Kernel, 3L} & label & 97.57$\pm$0.41 & \bf98.29 & \bf92.30$\pm$0.40 & \bf93.11\\
	\textsc{Our LSTM, 2L} & word & 97.28$\pm$0.36 & 97.57 & 90.66$\pm$0.55 & 91.53\\
	\textsc{Our LSTM, 2L} & full (decoupled) & 97.22$\pm$0.32 & 97.14 & 91.53$\pm$0.50 & 92.62\\
\hline

  \hline
\end{tabular}
\caption{\label{table_snips}Test set results on the Snips dataset. (*) indicates numbers reported by \cite{goo2018slot}}
\end{table*}

\begin{table*}
 \small
  \centering
  \begin{tabular}{|l|c|c|c|c|c|}
    \hline
    & & \multicolumn{4}{c|}{ATIS}\\
    \hline
    & & \multicolumn{2}{c|}{IC Acc} & \multicolumn{2}{c|}{SLR F1}\\
    Model & Recurrence & Mean & Max & Mean & Max  \\
    \hline
    LSTM+attn+gates \cite{goo2018slot} & full & 94.10 & - & 95.20 & -\\
    '18 Two LSTMs \cite{wang2018bi} & full & - & \bf 98.99 & - & \bf 96.89\\
    '18 self-attn+LSTM \cite{li2018self} & full & - & \bf 98.77 & - & \bf 96.52\\
    \hline
		\textsc{Our CNN, 5Kernel, 3L} & none & 97.04$\pm$0.62 & 97.98 & 94.84$\pm$0.22 & 94.95\\
		\textsc{Our CNN, 5Kernel, 3L} & label & \bf97.37$\pm$0.57 & \bf98.10 & \bf95.27$\pm$0.19 & \bf95.54\\
		\textsc{Our LSTM, 2L} & word & 96.84$\pm$0.49 & 97.65 & 95.13$\pm$0.29 & 95.41\\
		\textsc{Our LSTM, 2L} & full (decoupled) & 97.00$\pm$0.44 & 97.98 & 95.15$\pm$0.25 & 95.21\\
    \hline
  \end{tabular}
  \caption{\label{table_atis}Test set results on the ATIS dataset, compared to recent recurrent models.
														 }
\end{table*}

 \subsection{High-performing convolutional models}
The larger convolutional models provide very high accuracy while maintaining fast inference and training speeds.
In particular, our best CNN model, \textsc{cnn-5kernel-3l}, achieves 94.22 SL F1 on Snips, compared to the two-layer LSTM with label-recurrence, which achieves 93.88.
The model achieves this modest improvement with over 2x the inference speed, training in under 1/2 the time, and demonstrating even stronger results on the test sets, discussed below.

On ATIS, where utterances are longer but slot label spans are shorter, LSTMs outperform CNNs on the development sets.


\subsection{Issues with self-attention}
Our strongest self-attention model underperforms CNNs and LSTMs on both Snips and ATIS, with a maximum SL of 89.31 and 95.86 on the datasets, respectively.
Though self-attention models have seen success in complex tasks with lots of training data, we suggest in this study that they lack the inductive biases to perform well on these small datasets.

Relative position embeddings go a long way in improving self-attention models; adding them to a 1-layer attentional encoder improves ATIS and Snips SL by approximately 24 and 22 points, respectively.
We find that adding attention heads does not add considerably to the computational complexity of attention models, while increasing accuracy; thus in a speed-accuracy tradeoff, it is likely better to add heads rather than layers as each layer adds $O(n^2*d_x)$ additional computations.



\subsection{Word and label recurrence in LSTMs}
Our LSTM word contextualization modules show that with recurrent word context modeling, label-recurrence is less important. 
For instance, 2-layer LSTM achieves only .78 increase in SL with label recurrence over independent prediction.

\subsection{Best models compared to prior work}
We report test set results on Snips and ATIS in Tables~\ref{table_snips} and \ref{table_atis}.
Our best models from our validation study, \textsc{cnn-5kernel-3l} and \textsc{lstm-2l}, outperform the state-of-the-art on the Snips dataset, with label-recurrence proving crucial, especially for Snips.
In particular, \textsc{cnn-5kernel-3l} with label recurrence achieves an average SL F1 of 92.30, improving over the previous state-of-the-art of 88.8, by reducing error rate by \textbf{30\%},
 and .57-point improvement on IC.

On ATIS, our label-recurrent models outperform slot-gated LSTM model of \citet{goo2018slot} on both IC and SL tasks. 
\citet{wang2018bi} attribute their result to using IC and SL-specific LSTMs and use 300-dimensional word embedding and 200-dimensional LSTMs, but with an ATIS vocabulary of 867 words (suggesting a relatively simple sequence space), we are unable to determine the source of the improvement from a modeling standpoint.  Similar observation was made for \cite{li2018self} where 264-dimension embeddings is used. 
There is a high chance that these models might be severely overfitted. 

We hypothesize that our models perform better on Snips because much of Snips slot labeling depends on consistency within long spans, whereas ATIS slot labels have longer-distance dependencies, for example between \texttt{to\_city} and \texttt{from\_city} tags.

\begin{figure}
  \centering
  \includegraphics[width=.75\linewidth]{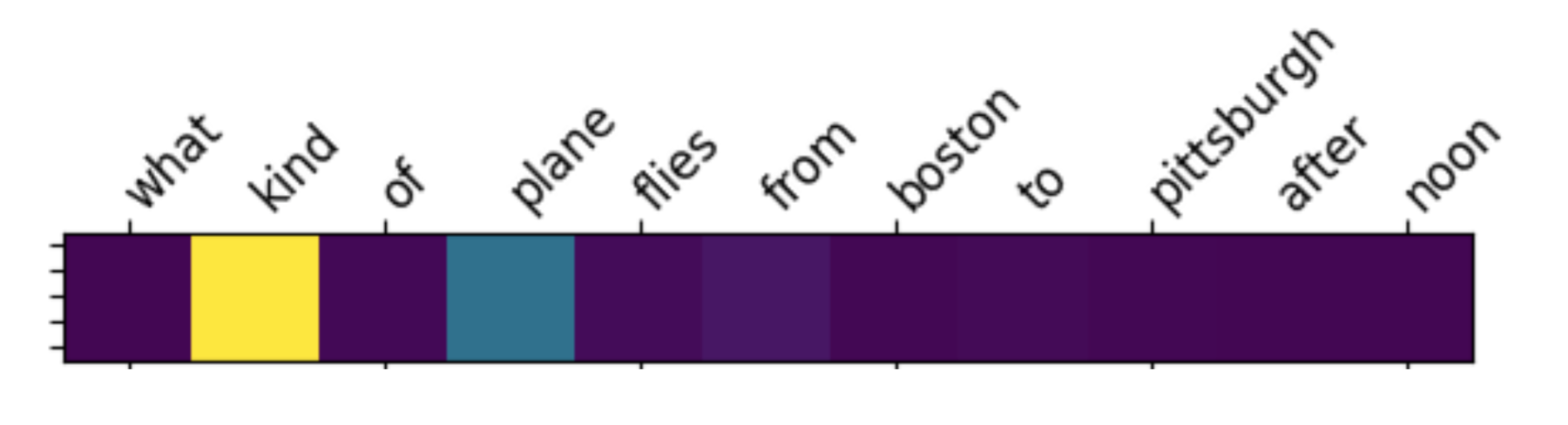}
  \caption{ \small Visualization of the weight given to each token representation by the attention-based pooling for sentence representation. Lighter colors indicate greater attention.}
  \label{attn_A}
\end{figure}

\subsection{Attention Visualization}
We note that anecdotally, few words in each utterance are useful in indicating the intent. In the example given in Figure \ref{attn_A}, presence of possible departure and arrival cities may be distracting, but the attention mechanism correctly learns to focus on words that indicate \texttt{atis\_aircraft} intent.


\section{Conclusion}


We presented a general family of joint IC+SL neural architectures that decomposes the task into modules for analysis. Using this framework, we conducted an extensive study of word contextualization methods (including utility of recurrence in the representation and output space) and determined that label-recurrent models, with non-recurrent word representation and a recurrent model of slot label dependencies, are a good fit for high performance in both accuracy and speed.

With the results of this study, we proposed a convolution-based joint IC+SL model for SLU that achieves new state-of-the-art results on Snips dataset while maintaining a simple design, shorter training, and faster inference than comparable recurrent methods.

\section{Implementation details}
For all models, we randomly initialize word embeddings and use $d_x = 70$.
We optimize using Adadelta algorithm \cite{zeiler2012adadelta}, with initial learning rate, $.01$.
We clip and pad all training and development sentences to length 30, with clipping affecting a small number of utterances.
Dropout \cite{srivastava2014dropout} probability of $.3$ is used in all models.
We train using a batch size of 128 split across 4 GPUs on a p3.8xlarge EC2 instance, and perform inference using CPUs on same machine.

\bibliography{acl2019}
\bibliographystyle{acl2019}

\end{document}